\documentclass[runningheads]{llncs}
\usepackage{graphicx}%
\usepackage{multirow}%
\usepackage{amsmath,amssymb,amsfonts}%
\usepackage{mathrsfs}%
\usepackage[title]{appendix}%
\usepackage{xcolor}%
\usepackage{textcomp}%
\usepackage{manyfoot}%
\usepackage{booktabs}%
\usepackage{algorithm}%
\usepackage{algorithmicx}%
\usepackage{algpseudocode}%
\usepackage{listings}%

\usepackage[justification=centering]{caption}
\usepackage{CJKutf8}
\usepackage{multicol}
\usepackage{bm}
\usepackage{makecell}
\usepackage{CJKutf8}
\usepackage{tcolorbox}
\usepackage[font=tiny]{subfig}
\usepackage{caption}
\usepackage{overpic}
\usepackage[misc]{ifsym}
\usepackage{hyperref}

\begin{document}
\title{scCDCG: Efficient Deep Structural Clustering for single-cell RNA-seq via Deep Cut-informed Graph Embedding}
\titlerunning{Deep Structural Clustering for single-cell RNA-seq data}
%

\author{Ping Xu\inst{1,2}$^*$ \and
Zhiyuan Ning\inst{1,2}$^*$ \and
Meng Xiao\inst{1,2} \and 
Guihai Feng\inst{4,5,6} \and 
Xin Li\inst{4,5,6} \and 
Yuanchun Zhou\inst{1,2,3} \and 
Pengfei Wang\inst{1,2}$^{(\textrm{\Letter})}$}
   
\institute{
Computer Network Information Center, Chinese Academy of Sciences\and
University of Chinese Academy of Sciences\and
Hangzhou Institute for Advanced Study, University of Chinese Academy of Sciences\and
State Key Laboratory of Stem Cell and Reproductive Biology, Institute of Zoology, Chinese Academy of Sciences\and 
Institute for Stem Cell and Regenerative Medicine, Chinese Academy of Science\and
Beijing Institute for Stem Cell and Regenerative Medicine, Chinese Academy of Science\\
\email{\{xuping,ningzhiyuan,shaow,zyc\}@cnic.cn, \{fenggh,xinli\}@ioz.ac.cn, pfwang@cnic.cn}
}
\authorrunning{P.Xu et al.}


%
\maketitle              

\def\thefootnote{*}\footnotetext{These authors contributed equally to this work.}

\vspace{-1cm}

\begin{abstract}
Single-cell RNA sequencing (scRNA-seq) is essential for unraveling cellular heterogeneity and diversity, offering invaluable insights for bioinformatics advancements. 
Despite its potential, traditional clustering methods in scRNA-seq data analysis often neglect the structural information embedded in gene expression profiles, crucial for understanding cellular correlations and dependencies.
Existing strategies, including graph neural networks, face challenges in handling the inefficiency due to scRNA-seq data's intrinsic high-dimension and high-sparsity.
Addressing these limitations, we introduce scCDCG (\textbf{s}ingle-\textbf{c}ell RNA-seq \textbf{C}lustering via \textbf{D}eep \textbf{C}ut-informed \textbf{G}raph), a novel framework designed for efficient and accurate clustering of scRNA-seq data that simultaneously utilizes intercellular high-order structural information. 
scCDCG comprises three main components:
(i) \textit{A graph embedding module utilizing deep cut-informed techniques}, which effectively captures intercellular high-order structural information, overcoming the over-smoothing and inefficiency issues prevalent in prior graph neural network methods.
(ii) \textit{A self-supervised learning module guided by optimal transport}, tailored to accommodate the unique complexities of scRNA-seq data, specifically its high-dimension and high-sparsity.
(iii) \textit{An autoencoder-based feature learning module} that simplifies model complexity through effective dimension reduction and feature extraction.
Our extensive experiments on 6 datasets demonstrate scCDCG's superior performance and efficiency compared to 7 established models, underscoring scCDCG's potential as a transformative tool in scRNA-seq data analysis.
Our code is available at: \url{https://github.com/XPgogogo/scCDCG}. 
\end{abstract}

\section{Introduction}
Single-cell RNA sequencing (scRNA-seq) data analysis is a groundbreaking advancement in the field of bioinformatics that enables researchers to explore the complex structure and heterogeneity within cell populations and reveal different cell types, states, or subpopulations~\cite{shapiro2013single,wang2025sccompass,yang2023genecompass}.
Clustering is an important tool for analyzing scRNA-seq data, which separates the cells into different groups with similar cells in the same group, thus can reveal the complex characteristics of different cell populations~\cite{kiselev2019challenges}.
Clustering of scRNA-seq data has proven to be particularly valuable in elucidating cell development, understanding disease mechanisms, and identifying novel cell types and subtypes~\cite{xu2025scsiameseclu}. 

Early classical scRNA-seq clustering methods rely on hard clustering algorithms.
For example, pcaReduce~\cite{vzurauskiene2016pcareduce} combines two methods, PCA and K-means clustering, 
and iteratively merges clusters based on the associated probability density function; 
SUSSC~\cite{wang2021suscc} combines Uniform Manifold Approximation and Projection (UMAP) with K-means clustering to merge clusters. 
However, the inherent characteristics of scRNA-seq data include high-dimension, high-sparsity, noise, nonlinearity, and frequent dropout events, which make the relationships between variables complex and nonlinear~\cite{eraslan2019single}. 
This will bring a common limitation to these methods: they use hard clustering algorithms that are difficult to deal with high-dimension and capture complex nonlinear data features, and are susceptible to interference from the inherent noise of the data.
Another line of research focuses on applying deep learning to the task of clustering scRNA-seq data. 
For example, DEC~\cite{xie2016unsupervised} uses neural networks to learn feature representations and cluster assignments; 
scDeepCluster~\cite{tian2019clustering} uses the ZINB model~\cite{eraslan2019single} to simulate the distribution of scRNA-seq data and combines it with deep learning embedding clustering.
However, all the above approaches focus on learning the features of a single cell while ignoring the intercellular structural information, which is important for describing the differences between cells.

To bridge this gap, a series of methods that create cell-cell graphs based on relationships between cells have emerged to incorporate intercellular structural information~\cite{gan2018identification}. 
Moreover, in order to introduce high-order structural information, the latest scRNA-seq data clustering methods have started to use Graph Neural Networks (GNNs). 
Specifically, scGNN~\cite{wang2021scgnn} utilizes GNNs and three multi-modal autoencoders to formulate and aggregate cell-cell relationships; 
scDSC~\cite{gan2022deep} combines a ZINB-based autoencoder and GNNs in a mutually supervised manner, enhancing the clustering performance. 
Despite considering intercellular structure information in scRNA-seq data, the GNN-based clustering still has some shortcomings:
(i) As single-cell transcriptome sequencing technology continues to advance, the scale of scRNA-seq data is getting increasingly larger. However, GNNs have high complexity and poor efficiency, i.e., longer training/inference time, especially when facing large-scale datasets~\cite{WANG2021165}. 
(ii) GNNs tend to generate similar representations for all nodes~\cite{ning2022graph,zhu2021graph,wang2024deep,ning2025rethinking}, which will even lead to the over-smoothing issue~\cite{chen2020measuring} (i.e., indistinguishable representations of nodes in different classes) when GNNs become deeper (i.e., more layers). 
This problem will become more severe when GNNs face the scRNA-seq data, because scRNA-seq data's high-dimension and high-sparsity will result in the similarity between all cell being close to 0, thus all cells are more poorly distinguishable.
Therefore, GNN-based clustering methods face challenges in handling the inefficiency due to scRNA-seq data’s intrinsic characteristics.

Thus, it is crucial to develop a scRNA-seq data clustering method that simultaneously utilizes intercellular high-order structural information, handles high-dimensional and high-sparse scRNA-seq datasets, and demonstrates high efficiency. 
In order to achieve this goal, we introduce \textbf{scCDCG} (\textbf{s}ingle-\textbf{c}ell RNA-seq \textbf{C}lustering via \textbf{D}eep \textbf{C}ut-informed \textbf{G}raph). 
The scCDCG consists of three main modules, specifically: 
(i) scCDCG incorporates complex intercellular high-order structural information through a novel and simple method of graph embedding module based on deep cut-informed techniques~\cite{shi2000normalized,ning2025deep}, while avoiding the efficiency limitations and over-smoothing issue of GNN-based methods~\cite{ning2021lightcake}. 
It is well-suited for high-dimensional and high-sparse scRNA-seq data, demonstrating enhanced efficiency. 
(ii) scCDCG utilizes a new self-supervised learning module based on optimal transport to guide the model learning process~\cite{monge1781memoire}, which is more adaptive to high-dimensional and high-sparse scRNA-seq data and enhances clustering accuracy and robustness. 
(iii) scCDCG employs an autoencoder-based feature learning module for dimension reduction and feature extraction, effectively reducing the model complexity. 
To fully verify the capacity and effectiveness of scCDCG, we conduct extensive experiments with 7 established models on 6 datasets, the results demonstrate the superior clustering performance and efficiency of scCDCG. 
In addition, we conduct sufficient ablation and analysis experiments to validate the effectiveness of each module of scCDCG. 

Our key contributions can be summarized as follows:
\begin{itemize}
    \item We propose scCDCG, modeled from a graph cutting perspective, which is more suitable for high-dimensional, high-sparse scRNA-seq data, exploiting the complex intercellular higher-order structural information while avoiding the over-smoothing problem and efficiency limitations of previous GNN-based methods. 
    \item We introduce a self-supervised learning module via optimal transport to guide the model learning process, which can better adapt to the complex structure of scRNA-seq data, especially the properties of high-dimension and high-sparsity.
    \item Extensive experiments are conducted on various real-world scRNA-seq datasets to demonstrate the excellent efficiency and superior performance of scCDCG. We also perform comprehensive ablation experiments to demonstrate that each component of our method is indispensable.
\end{itemize} 
\section{Related Work}
\subsection{Classical Clustering Methods for ScRNA-seq Data}
In scRNA-seq data analysis, early clustering methods primarily employed hard clustering algorithms. pcaReduce~\cite{vzurauskiene2016pcareduce} is a notable example, combining Principal Component Analysis (PCA) with K-means clustering. This method iteratively merges clusters based on probability density functions, capturing key data variances.
Similarly, SUSSC~\cite{wang2021suscc} integrates Uniform Manifold Approximation and Projection (UMAP) with K-means, enhancing data dimensionality reduction and cluster accuracy.
Deep learning also gained prominence in scRNA-seq clustering. DEC~\cite{xie2016unsupervised} employs neural networks for feature representation and clustering, adapting to complex gene expression profiles. Meanwhile, scDeepCluster~\cite{tian2019clustering} utilizes the Zero-Inflated Negative Binomial (ZINB) model~\cite{eraslan2019single} in conjunction with deep learning, addressing data sparsity and over-dispersion for more precise clustering.
However, all the above approaches focus on learning the features of a single cell while ignoring the intercellular structural information, which is important for describing the differences between cells.

\subsection{Deep Structural Clustering Methods for ScRNA-seq Data} 
In the field of single-cell RNA sequencing (scRNA-seq) data analysis, deep structural clustering methods have gained prominence. Notably, scGNN employs Graph Neural Networks (GNNs)~\cite{wang2021scgnn} and multi-modal autoencoders to aggregate cell-cell relationships and model gene expression patterns using a left-truncated mixture Gaussian model. scDSC~\cite{gan2022deep}, on the other hand, combines a ZINB model-based autoencoder with GNN modules, integrating these using a mutually supervised strategy for enhanced data representation. GraphSCC~\cite{zeng2020accurately} utilizes a graph convolutional network to capture structural cell-cell relationships, optimizing the network's output with a self-supervised module. Lastly, scGAE~\cite{luo2021topology} leverages a graph autoencoder to maintain both feature and topological structure information of scRNA-seq data, illustrating the diversity and sophistication of current approaches in the field.

\section{Methodology}
\subsection{Definition}
Assume the scRNA-seq data as $\bm{X} \in \mathbb{R}^{n\times m}$, where $\bm{x}_{ij}\left(1\leq i\leq n, 1\leq j \leq m\right)$ shows $j$-th gene’s expression in the $i$-th cell.
Denote $\bm{H}_{n\times d}=[\bm{h}_1, \bm{h}_2, \cdots, \bm{h}_n]^{T}$ is the latent embedding of $\bm{X}$ and $\bm{c} = [c_{1},...,c_{n}]$ is the clustering assignment. We will simultaneously learn the embedding $\bm{H}$ and clustering assignments $\bm{c}$ in this paper.\par
\begin{figure*}[t!]
    \centering
    \vspace{-2mm}
    \includegraphics[width=1\textwidth]{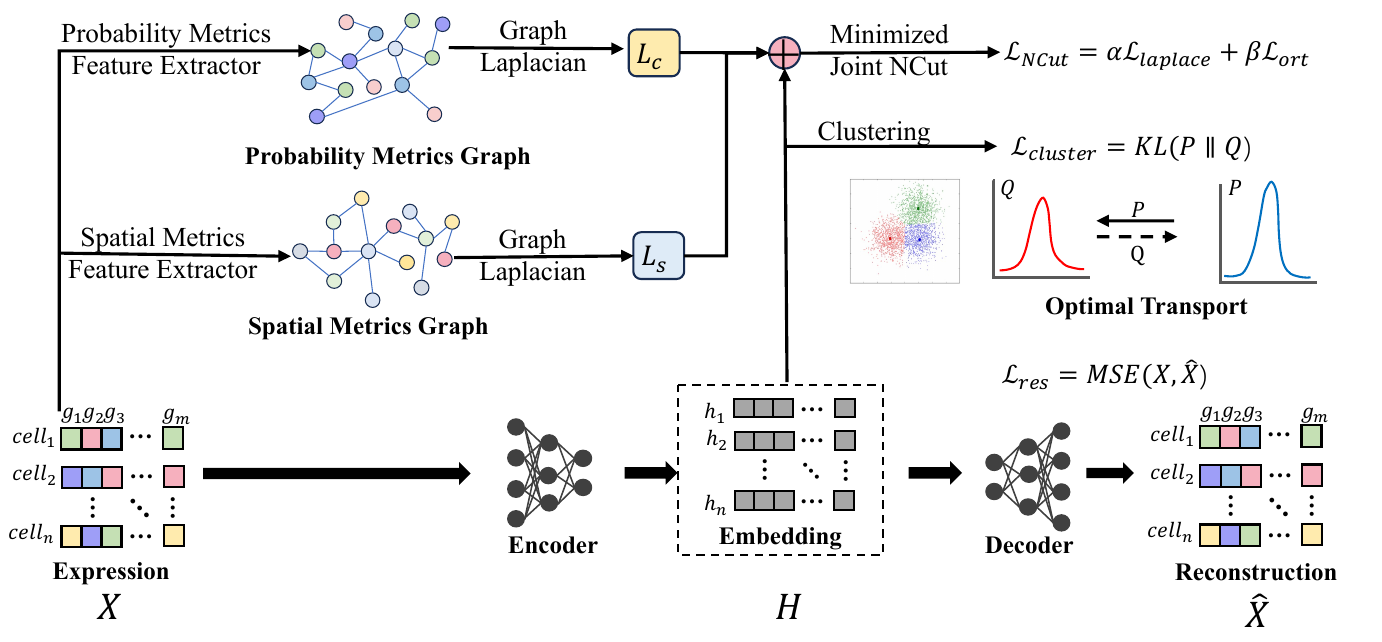}
    \vspace{-2mm}
    \caption{The model architecture of scCDCG. The scCDCG contains three core modules: (i) an autoencoder, (ii) a two-channel graph embedding module, and (iii) an optimal transport-based self-supervision learning module. 
    }
    \label{fig:overview}
    \vspace{-4mm}
\end{figure*} 
\subsection{The framework of scCDCG}
We propose scCDCG, a deep cut-informed graph model-based single cell clustering method (see Fig.~\ref{fig:overview} for its architecture), which includes (i) an autoencoder-based feature learning module for learning gene expression embeddings, (ii) a graph embedding module based on deep cut-informed techniques for capturing intercellular high-order structural information, and (iii) a self-supervision learning module via optimal transport for generating clustering assignments. 
\subsubsection{Feature learning module based an autoencoder.} To reduce complexity (i.e., using the mean square error loss function), which can map high-dimensional scRNA-seq data to a low-dimensional feature space. We adopt an unsupervised autoencoder~\cite{hinton2006reducing}. 
The encoder of the autoencoder maps $\bm{X}$ to the latent representation $H$ as:
\begin{equation}
    \bm{H} = f_{enc}(W\bm{X} + b).
\label{equ:mlp_encoder}
\end{equation}
The decoder maps $H$ to a reconstruction $\bm{\hat{X}}$ of the raw data, as: 
\begin{equation}
    \bm{\hat{X}} = f_{dec}(W^{'} \bm{H} + b^{'} ),
\label{equ:mlp_decoder}
\end{equation}
where $b$ and $b^{'}$, respectively, represent the bias vectors of the encoder and the decoder, $W$ and $W^{'}$ are the weight matrix of the encoder and the decoder.\par
Thus, the reconstruction loss function can be written as:
\begin{equation}
    \mathcal{L}_{res}=\frac{1}{2N}\sum_{i=1}^{N}\|x_i-\hat{x_i}\|_2 ^ 2=\frac{1}{2N}\|X-\hat{X}\|_F ^2.
\end{equation}

\subsubsection{Graph embedding module based on deep cut-informed techniques.} This module can leverage intercellular high-order structural information and avoid the over-smoothing issue and scalability limitations of GNN-based methods, making it well-suited for high-dimension and high-sparsity scRNA-seq data and demonstrating enhanced scalability.

Specifically, we employ a dual-channel approach, processing the raw data $X$ concurrently through both probability metrics and spatial metrics feature extractors, which yields the probability metrics graph $\mathcal{G_C}$ and the spatial metrics graph $\mathcal{G_S}$. Here, $\mathcal{V}$ is a set of $n$ nodes in the graph, $\mathcal{E} \subseteq \mathcal{V} \times \mathcal{V}$ is the edge set of graph and $\bm{A}\in\{0, 1\}^{n \times n}$ is the adjacency matrix of graph. The two channels are as follows:

\emph{Channel 1}: The probability metrics feature matrix $\bm{C} \in \mathbb{R}^{n \times n}$ is calculated as $\bm{C} = \bm{X}\bm{X}^T$. Each entry $\bm{C}_{ij}$ in the matrix represents the covariance between cells $i$ and $j$. Then we use the probability metrics matrix to construct a probability metrics graph $\mathcal{G_C}=(\mathcal{V}, \bm{C}, \mathcal{E}, \bm{A}_C)$, each node represents a cell.

\emph{Channel 2}: The spatial metrics feature matrix $\bm{S} \in \mathbb{R}^{n \times n}$is computed as $\bm{S}_{ij} = \frac{\bm{x}_i^T \bm{x}_j}{|\bm{x}_i|_2 |\bm{x}_j|_2}$. Each entry $\bm{S}_{ij}$ in the matrix represents the pairwise cosine similarity between cells $i$ and $j$. Then we utilize it to obtain a spatial metrics graph $\mathcal{G_S}=(\mathcal{V}, \bm{S}, \mathcal{E}, \bm{A}_S)$. 

Meanwhile, we proposed a graph embedding module based on deep cut-informed to fuse the probability metrics graph and spatial metrics graph by minimizing their joint normalized cut~\cite{shi2000normalized}. 
For graphs $\mathcal{G_C}$ and $\mathcal{G_S}$, $\bm{D}_{\bm{C}}$, $\bm{D}_{\bm{S}}$ are their diagonal matrices respectively. Here, the normalized graph Laplacian of $\bm{C}$ and $\bm{S}$ can be formulated as:
\begin{equation}
\begin{aligned}
      & L_{\bm{C}} = \bm{I} - \bm{D}_{\bm{C}}^{-1/2}\bm{C}\bm{D}_{\bm{C}}^{-1/2} ,\\
      & L_{\bm{S}} = \bm{I} - \bm{D}_{\bm{S}}^{-1/2}\bm{S}\bm{D}_{\bm{S}}^{-1/2} .
\label{equ:laplace}
\end{aligned}
\end{equation}
Given two subsets of $\mathcal{V}$ denoted as $\mathcal{V}_{1}$ and $\mathcal{V}_{2}$, we define $\bm{A}(\mathcal{V}_{1}, \mathcal{V}_{2}) = \sum_{i\in \mathcal{V}_{1}, j\in \mathcal{V}_{2}}A_{ij}$. Then for a partition $\mathcal{V} = \mathcal{V}_{1}\bigcup \cdots \bigcup \mathcal{V}_{K}$, its normalized cut~\cite{shi2000normalized} can be defined as follows:
\begin{equation}
\text{Ncut}(\mathcal{V}_{1},\cdots,\mathcal{V}_{K}) = \frac{1}{2}\sum_{i=1}^{K}\frac{\bm{A}(\mathcal{V}_{i}, \bar{\mathcal{V}}_{i})}{\text{vol}(\mathcal{V}_{i})},
\end{equation}
where $\bar{\mathcal{V}}_{i}$ is the complementary set of $\mathcal{V}_{i}$ and $\text{vol}(\mathcal{V}_{i}) = \bm{A}(\mathcal{V}_{i}, \mathcal{V}_{i}) + \bm{A}(\mathcal{V}_{i}, \bar{\mathcal{V}}_{i})$ is the total edges of $\mathcal{V}_{i}$. According to this criterion, an optimal graph partition should have minimized Ncut. However, it is NP-hard to find the optimal graph partition. Inspired by covariates-assisted spectral clustering, we can alternatively solve it with the following optimization strategy: 
\begin{equation}
\begin{aligned}
    & \arg\min_{\bm{H}} \text{Tr}(\bm{H}^{T}L_{\bm{C}}\bm{H}) & \text{subject to } \bm{H}^{T}\bm{H} = \bm{I},\\
    & \arg\min_{\bm{H}} \text{Tr}(\bm{H}^{T}L_{\bm{S}}\bm{H}) & \text{subject to } \bm{H}^{T}\bm{H} = \bm{I},
\label{equ:ncut}
\end{aligned}
\end{equation}
which is a relaxation of minimizing the normalized cut. 
$\bm{H}$ is a soft partition of graph $\mathcal{G}$ in terms of minimizing normalized cut, which implies that $\bm{H}$ can be viewed as the node embeddings for clustering. 
It is a natural idea to make the learned embedding $\bm{H}$ by Equation~\ref{equ:ncut} contain the information of both probability metrics and spatial metrics graphs, so the optimization function is following:
\begin{equation}
\begin{aligned}
    & \arg\min_{\bm{H}} \text{Tr}(\bm{H}^{T}(\alpha L_{\bm{C}} + (1-\alpha)L_{\bm{S}})\bm{H}) \\
    & \text{subject to } \bm{H}^{T}\bm{H} = \bm{I} ,
\label{equ:ncut optimization function}
\end{aligned}
\end{equation}
where $\alpha$ is the balance parameter between probability metrics graph and spatial metrics graph. 
So the NCut loss function can be written as:
\begin{equation}
    \mathcal{L}_{\text{NCut}} = \beta \text{Tr}(\bm{H}^{T}(\alpha L_{\mathcal{G}} + (1-\alpha)L_{\bm{S}})\bm{H}) + \gamma \|\bm{H}^{T}\bm{H} - \bm{I}\|_{F},
\label{equ:loss_ncut}
\end{equation}
where $\beta$, $\gamma$ are both tuning parameters and $\|\bm{H}^{T}\bm{H} - \bm{I}\|_{F}$ is the loss to guarantee the orthogonality. 

Note that, the learned embedding $\bm{H}$ is better for clustering from two viewpoints: (i) The loss function is a relaxation of minimizing the joint normalized cut of probability metrics graph and spatial metrics graph, which means $\bm{H}$ is a soft assignment of the graph partition from the perspective of minimizing normalized cut. (ii) The trade-off between the probability metrics graph and spatial metrics graph makes $\bm{H}$ consistent with both two graphs, which maintains the structure supported by both two graphs. 

\subsubsection{Self-supervision learning module via optimal transport.} To better adapt to the complex structure of scRNA-seq data, especially the two properties of high-dimension and high-sparsity, we adopt a self-supervised approach introduced by~\cite{xie2016unsupervised}, which can generate learning assignments. It employs the Student’s t-distribution~\cite{van2008visualizing} as a kernel to measure the similarity between the embedded point $h_{i}$ and the clustering center $c_{j}$. 
\begin{equation}
    q_{ij} = \frac{(1 + \|h_{i} - c_{j}\|^{2}/\theta)^{-\frac{1 + \theta}{2}}}{\sum_{j^{'}}(1 + \|h_{i} - c_{j^{'}}\|^{2}/\theta)^{-\frac{1 + \theta}{2}}},
\label{equ:kl_qij}
\end{equation}
where $h_{i}=f(\bm{x}_{i}) \in \bm{H}$ corresponds to $\bm{x}_{i} \in \bm{X}$ after embedding, $\theta$ is the degrees of freedom of the Student’s t-distribution, and $q_{ij}$ can be interpreted as the probability of assigning sample $i$ to cluster $j$ (i.e., a soft assignment). 
Then, we define $Q=\left[q_{i j}\right]$ as the distribution of the assignments of all samples.\par
After obtaining the clustering result distribution $Q$, we aim to optimize the data representation by learning the high-confidence assignments~\cite{ning2024fedgcs}. 
This approach aims to minimize the difference between the data representations and the cluster centers, thereby improving the cohesion of the clusters. For instance, SDCN~\cite{bo2020structural} calculated a target distribution $P=\left[p_{i j}\right]$:
\vspace{-2mm}
\begin{equation}
    p_{ij}=\frac{q_{ij} ^2 / f_j}{\sum{q_{ij^{'}}^2 / f_{j^{'}}}},
\end{equation}
where $f_j=\sum_{i} a_{ij}$ are soft cluster frequencies. In the target distribution $P$, each assignment in $Q$ is squared and normalized to enhance assignment confidence~\cite{bo2020structural}. However, to prevent the emergence of degenerate solutions that assign all data points to a single (arbitrary) label, we introduce constraints that align the label distribution with the mixing proportions. This approach ensures improved clustering accuracy and the equalized contribution of each data point to the loss calculation. Therefore, we construct the target probability matrix $P$ by solving the following optimal transport strategy:
\begin{equation}
\vspace{-2mm}
\begin{aligned}
        \min_{P} \quad & -P*(\text{log}Q) \\
        \mbox{s.t.}\quad & P\in\mathbb{R}_{+}^{N\times C}, \\
         \quad &P\ \bm{1}_C = \bm{1}_N \ \text{ and } \ P^{T}\bm{1}_N = N\bm{\pi}.
\end{aligned}
\label{equ:first_ot_optimization function}
\end{equation}
Here, we consider the target distribution $P$ as the transport plan matrix within the optimal transport theory and $- \text{log}Q$ as the associated cost matrix. The constraint $P^{T}\bm{1}_N = N\bm{\pi}$ is imposed, with $\pi$ representing the proportion of points in each cluster, which can be estimated by the intermediate clustering result. This step successfully adds the constraint that the result cluster distribution must be consistent with the mixing proportions. Given the computational burden of direct optimization, we leverage the Sinkhorn distance~\cite{sinkhorn1967diagonal} for rapid optimization through an entropic constraint. The optimization problem with an entropy constraint Lagrange multiplier is as follows:
\begin{equation}
\vspace{-2mm}
\begin{aligned}
        \min_{P} \quad & -P*(\text{log}Q)-\frac{1}{\lambda}\text{H}(P) \\
        \mbox{s.t.}\quad & P\in\mathbb{R}_{+}^{N\times C}, \\
         \quad &P\ \bm{1}_C = \bm{1}_N \ \text{ and } \ P^{T}\bm{1}_N = N\bm{\pi},
\end{aligned}
\label{equ:ot_optimization function}
\end{equation}
where $\bm{H}$ is the entropy function and $\lambda$ is the smoothness parameter controlling the equilibrium of clusters. The existence and unicity of $P$ are guaranteed, and can be effciently solved by Sinkhorn's~\cite{sinkhorn1967diagonal} fixed point iteration:
\begin{equation}
\vspace{-2mm}
\hat{P}^{(t)}=\operatorname{diag}\left(\boldsymbol{u}^{(t)}\right) Q^\lambda \operatorname{diag}\left(\boldsymbol{v}^{(t)}\right)
\end{equation}
where $t$ denotes the iteration and in each iteration: $\bm{u}^{(t)}=\bm{1}_N / (Q^{\lambda}\bm{v}^{(t-1)})$ and $\bm{v}=N_\pi / (Q^{\lambda}\bm{u}^{(t)})$, with the initiation $\bm{v}^{(0)}=\bm{1}_N$. After the fixed point iteration, we can get the optimal transport plan matrix $\hat{P}$, i.e., the solution of (\ref{equ:loss_kl}).\par
During the training process, we fix $\hat{P}$ and make $Q$ be consistent with $\hat{P}$, so the clustering loss function can be formulated as:
\begin{equation}
    \mathcal{L}_{\text{KL}}=\bm{\text{KL}}(\hat{P} \| Q).
\label{equ:loss_kl}
\end{equation}
Thus, the overall loss function is 
\begin{equation}
    \mathcal{L}=\mu\mathcal{L}_{res}+\sigma\mathcal{L}_{NCut}+\tau\mathcal{L}_{KL},
\label{equ:total_loss}
\end{equation}
where $\mu$, $\sigma$ and $\tau$ are tuning parameters to balance the three kinds of loss, $\mathcal{L}_{res}$, $\mathcal{L}_{NCut}$ and $\mathcal{L}_{KL}$. \par
\subsection{Overall algorithm}
In practice, we first train the model with both NCut loss and reconstruction loss by 200 epochs. After that, we apply K-means to the embeddings to obtain the initial centroids and clustering sizes. Then, we train the model with 200 epochs each for NCut loss, reconstruction loss, and clustering loss. Ultimately, clustering results are obtained at the last epoch. 

\section{Results}
\subsection{Experimental Setup}
\begin{table}[t!]
    \centering
    \vspace{-3mm}
    \caption{Summary of the scRNA-seq datasets.}
    \small
    \begin{tabular}{lccc}
         \toprule
         \textbf{Datasets} & \textbf{\makecell{Sample\\ Size}} & \textbf{\makecell{Gene\\ Number}}& \textbf{\makecell{Group\\ Number}}\\
         \midrule
         Human Pancreas cells 2~\cite{baron2016single} & 1724 & 20125 & 14\\
         Meuro Human Pancreas cells~\cite{muraro2016single} & 2122 & 19046 & 9\\
         Mouse Bladder cells~\cite{han2018mapping} & 2746 & 20670 & 16\\
         Worm Neuron cells~\cite{cao2017comprehensive} & 4186 & 13488 & 10\\
         CITE-CMBC~\cite{han2018mapping} & 8617 & 2000 & 15\\
         Human Liver cells~\cite{macparland2018single} & 8444 & 4999 & 11\\
        \bottomrule
    \end{tabular}
    \vspace{-3mm}
    \label{tab:dataset}
\end{table}
\subsubsection{Dataset.} 
We assessed the performance of the model using six scRNA-seq datasets, each of which was accompanied by provided cell labels. 
Of the six datasets, four are derived from human samples, one from mouse samples, and one from worm samples. The datasets cover a wide range of cell types, including pancreas, bladder, neurons, liver, and peripheral blood mononuclear cells. A detailed description of these datasets is provided in Table~\ref{tab:dataset}.
\subsubsection{Evaluation metrics.} 
To show the effectiveness of the proposed method, the clustering performance is evaluated by three established metrics from the public domain: Accuracy (ACC), Normalized Mutual Information (NMI)~\cite{strehl2002cluster}, and Adjusted Rand Index (ARI)~\cite{vinh2009information}. 
\subsubsection{Competing Methods.} 
We compare our model with the 7 competing methods to further evaluate its clustering performance. Detailed categorization and description are as follows:
\begin{itemize}
    \item \textbf{Early Clustering Methods}: \textbf{pcaReduce}~\cite{vzurauskiene2016pcareduce} iteratively combines clusters based on relevant probability density functions; \textbf{SUSSC}~\cite{wang2021suscc} uses stochastic gradient descent and nearest neighbor search for clustering. 
    \item \textbf{Deep Clustering Methods}: \textbf{DEC}~\cite{xie2016unsupervised} is deep-embedded clustering, employing neural networks to learn both feature representations and cluster assignments; \textbf{scDeepCluster}~\cite{tian2019clustering} utilizes the ZINB model to simulate the distribution of scRNA-seq data and combines deep embedding clustering to learn feature representation and clustering simultaneously; \textbf{scNAME}~\cite{wan2022scname} integrates a mask estimation task and a neighborhood contrastive learning framework for denoising and clustering scRNA-seq data.
    \item \textbf{Deep Structural Clustering Methods}: \textbf{scGNN}~\cite{wang2021scgnn} utilizes GNNs and three multi-modal autoencoders to formulate and aggregate cell-cell relationships and models heterogeneous gene expression patterns using a left-truncated mixture Gaussian model; \textbf{scDSC}~\cite{gan2022deep} learns data representations using the ZINB model-based autoencoder and GNN modules, and uses mutual supervised strategy to unify these two architectures.
\end{itemize}
\subsubsection{Implementation Details.}
In the experiments, we implement scCDCG in Python 3.7 based on PyTorch. For baseline approaches, we report the results stated in their papers. 
The training phase uses the same hyperparameters as the pre-training phase, such as learning rate, weight decay, and other parameters. 
The encoder network contains layers of $[256, 16]$ to produce a latent representation of 16 dimensions, and the decoder network has a symmetric structure. 
The smoothness parameter $\lambda$ of the optimal transport strategy is set as $5$ for all datasets; the degrees of freedom of the Student’s t-distribution $\theta$ is set as $1$ for all datasets; besides that, each dataset has unique hyperparameters, which are achieved by the grid search. 
The network is trained by the Adam optimizer with the initial learning rate and weight decay to update variables. 
Furthermore, hyperparameter settings for all datasets are available in the code. During network training, we train the scCDCG for 200 epochs. 
In order to ensure the accuracy of the experimental results, we conducted each experiment 10 times with random seeds on all the datasets and reported the mean and variance of the results.\par
\subsection{Overall Clustering Performance}
\begin{table*}[t!]
    \centering
    \vspace{-5mm}
    \caption{Clustering performances of all models on ten datasets (mean $\pm$ std). The best results are highlighted in \textbf{bold}, and the runner-up results are highlighted in \underline{underline}. (Higher values indicate better performance.)}
    \resizebox{\textwidth}{!}
    { 
    \renewcommand\arraystretch{1.7}
    \begin{tabular}{c|c|ccccccc|c}
        \toprule
        \textbf{\large Dataset} & \textbf{\large Metric} & \textbf{\large \makecell{\large pcaReduce}} & \textbf{\large \makecell{SUSSC}} & \textbf{\large \makecell{DEC}} & \textbf{\large \makecell{scDeep-\\Cluster}} & \textbf{\large scNAME} & \textbf{\large scGNN} & \textbf{\large scDSC} & \textbf{\large \makecell{scCDCG}}\\
        \midrule
        
        \multirow{3}*{\textbf{\large \makecell{Human\\ Pancreas\\ cells 2}}}
        & \large ACC & \large 37.56 \small$\pm$ 0.4 & \large 46.75 \small$\pm$ 0.2 & \large 45.26 \small$\pm$ 1.8 &\large62.88 \small$\pm$ 5.5 &\large 63.33 \small$\pm$ 2.0 & \large 57.62 \small$\pm$ 2.3 &\large \underline{80.40} \small$\pm$ 4.0 &\large \textbf{84.66} \small$\pm$ 1.1\\
        & \large NMI & \large 50.81 \small$\pm$ 0.3 & \large 68.62 \small$\pm$ 0.0 & \large66.32 \small$\pm$ 2.1 &\large 73.97 \small$\pm$ 2.2 &\large 58.05 \small$\pm$ 1.6 & \large 79.32 \small$\pm$ 2.3 &\large \underline{79.44} \small$\pm$ 1.7 &\large \textbf{83.44} \small$\pm$ 2.6\\
        & \large ARI & \large 19.23 \small$\pm$ 0.5 & \large 38.45 \small$\pm$ 0.4 & \large39.14 \small$\pm$ 1.8 &\large 64.56 \small$\pm$ 7.5 & \large 32.12 \small$\pm$ 1.5 & \large 79.96 \small$\pm$ 1.7 &\large \underline{82.85} \small$\pm$ 6.5 &\large \textbf{85.30} \small$\pm$ 1.6\\
        \hline
        
        \multirow{3}*{\textbf{\large \makecell{Meuro \\Human\\ Pancreas \\ cells}}}
        & \large ACC & \large 50.56 \small$\pm$ 3.4 & \large 76.26 \small$\pm$ 0.1 &\large 72.22 \small$\pm$ 5.5 &\large 60.89 \small$\pm$ 3.1 &\large \underline{80.36} \small$\pm$ 7.9 & \large 75.12 \small$\pm$ 4.2 &\large 69.18 \small$\pm$ 3.7 &\large \textbf{92.65} \small$\pm$ 1.9\\
        & \large NMI & \large 54.83 \small$\pm$ 1.7 & \large 62.25 \small$\pm$ 0.0 & \large76.29 \small$\pm$ 2.7 &\large 70.02 \small$\pm$ 2.0 & \large \underline{79.12} \small$\pm$ 2.7 & \large 72.15 \small$\pm$ 2.2 & \large 64.31 \small$\pm$ 1.3 &\large \textbf{86.81} \small$\pm$ 1.0\\
        & \large ARI & \large 36.61 \small$\pm$ 2.3 & \large 64.23 \small$\pm$ 0.0 & \large61.76 \small$\pm$ 5.6 & \large 55.14 \small$\pm$ 2.4 &\large \underline{76.6} \small$\pm$ 10.35 & \large 72.13 \small$\pm$ 3.1 & \large 56.63 \small$\pm$ 1.2 & \large \textbf{91.37} \small$\pm$ 1.2\\
        \hline

        \multirow{3}*{\textbf{\large \makecell{Mouse \\Bladder\\ cells}}}
        & \large ACC & \large 31.75 \small$\pm$ 0.5 & \large 50.60 \small$\pm$ 1.6 &\large 50.72 \small$\pm$ 2.9 &\large 74.30 \small$\pm$3.8 &\large 62.45 \small$\pm$ 4.0 & \large 54.21 \small$\pm$ 3.2 &\large \underline{74.32} \small$\pm$ 3.9 &\large \textbf{75.61} \small$\pm$ 1.2\\
        & \large NMI & \large 44.15 \small$\pm$ 0.3 & \large 65.35 \small$\pm$ 0.1 & \large 62.52 \small$\pm$ 2.2 &\large \underline{74.99} \small$\pm$ 2.0 & \large 61.96 \small$\pm$ 1.9 & \large 69.86 \small$\pm$ 1.7 &\large 74.42 \small$\pm$ 2.4 & \large \textbf{75.91} \small$\pm$ 0.9\\
        & \large ARI & \large 36.61 \small$\pm$ 2.3 & \large 39.35 \small$\pm$ 1.0 & \large41.86 \small$\pm$ 6.28 & \large 62.33 \small$\pm$ 5.4 & \large 39.51 \small$\pm$ 1.6 & \large 56.45 \small$\pm$ 5.3 & \large \underline{62.52} \small$\pm$ 4.7 &\large \textbf{64.55} \small$\pm$ 2.3\\
        \hline
        
        \multirow{3}*{\textbf{\large \makecell{Worm \\Neuron\\ cells}}}
        & \large ACC & \large 29.30 \small$\pm$ 3.7 & \large 54.94 \small$\pm$ 0.8 & \large 40.96 \small$\pm$ 3.9 & \large65.71 \small$\pm$ 4.0 & \large \underline{66.89} \small$\pm$ 2.0 & \large 59.32 \small$\pm$ 2.1 & \large 60.93 \small$\pm$ 3.7 & \large \textbf{78.73} \small$\pm$ 2.9\\
        & \large NMI & \large 25.45 \small$\pm$ 14.4 & \large 43.84 \small$\pm$ 0.1 & \large38.61 \small$\pm$ 2.5 & \large \underline{67.58} \small$\pm$ 1.1 & \large 64.23 \small$\pm$ 2.1 & \large 48.73 \small$\pm$ 1.4 &\large 61.63 \small$\pm$ 3.7 & \large \textbf{72.11} \small$\pm$ 2.0\\
        & \large ARI & \large 8.65 \small$\pm$ 8.8 & \large 32.69 \small$\pm$ 5.5 & \large19.99 \small$\pm$  1.5 &\large 49.91 \small$\pm$ 4.3 & \large \underline{55.59} \small$\pm$ 1.1 & \large 30.68 \small$\pm$ 1.8 &\large 54.10 \small$\pm$ 4.2 & \large \textbf{59.78} \small$\pm$ 1.9\\
        \hline

        \multirow{3}*{\textbf{\large \makecell{CITE-\\CMBC}}}
        & \large ACC & \large 28.19 \small$\pm$ 0.1 & \large 47.22 \small$\pm$ 0.1 &\large 41.88 \small$\pm$ 5.4 &\large \underline{70.80} \small$\pm$ 2.5 & \large 68.06 \small$\pm$ 13.7 & \large 65.32 \small$\pm$ 2.0 &\large 67.69 \small$\pm$ 2.8 &\large \textbf{71.45} \small$\pm$ 1.8\\
        & \large NMI & \large 28.36 \small$\pm$ 0.2 & \large 60.14 \small$\pm$ 0.0 & \large46.87 \small$\pm$ 9.2 &\large \underline{72.53} \small$\pm$ 0.7 & \large 63.23 \small$\pm$ 13.2 & \large 61.42 \small$\pm$ 3.1 &\large 63.80 \small$\pm$ 2.7 & \large \textbf{74.77} \small$\pm$ 1.7\\
        & \large ARI & \large 5.10 \small$\pm$ 0.1 & \large 35.56 \small$\pm$ 1.6 & \large23.55 \small$\pm$ 10.3 & \large 56.23 \small$\pm$ 4.1 & \large \underline{58.99} \small$\pm$ 21.3 & \large 49.56 \small$\pm$ 1.6 &\large 50.09 \small$\pm$ 1.6 & \large \textbf{61.46} \small$\pm$ 1.4\\
        \hline

        \multirow{3}*{\textbf{\large \makecell{Human\\ Liver\\ cells}}}
        & \large ACC & \large 34.92 \small$\pm$ 0.2 & \large 46.62 \small$\pm$ 1.6 & \large 46.32 \small$\pm$ 5.5 & \large 63.44 \small$\pm$ 0.7 & \large \underline{73.55} \small$\pm$ 4.8 & \large 68.65 \small$\pm$ 3.2 & \large 67.90 \small$\pm$ 3.8 & \large \textbf{75.34} \small$\pm$ 1.7\\
        & \large NMI & \large 32.07 \small$\pm$ 0.6 & \large 67.32 \small$\pm$ 0.0 & \large52.24 \small$\pm$ 5.0 &\large 74.60 \small$\pm$ 2.5 & \large \underline{75.99} \small$\pm$ 2.0 & \large 62.33 \small$\pm$ 2.1 &\large 57.93 \small$\pm$ 2.6 & \large \textbf{79.34} \small$\pm$ 2.6\\
        & \large ARI & \large 6.34 \small$\pm$ 14.4 & \large 35.12 \small$\pm$ 0.0 & \large31.04 \small$\pm$ 4.7 & \large 58.58 \small$\pm$ 10.1 & \large \underline{72.40} \small$\pm$ 9.4 & \large 65.41 \small$\pm$ 1.3 & \large 57.34 \small$\pm$ 2.7 & \large \textbf{81.26} \small$\pm$ 2.7\\
        \bottomrule
    \end{tabular}
     }
    \label{tab:expriment}
\end{table*}

To verify the superior performance of scCDCG, we benchmarked it against 7 competing methods on 6 real-world datasets. In the comparative analysis, we utilize three widely used metrics (ACC, NMI, and ARI) to evaluate the clustering performance of each method, as summarized in Table~\ref{tab:expriment}. \par
From these results, we note several observations: 
For each metric, our approach scCDCG is significantly better than the best results of competing methods on all datasets, achieving a significant improvement of 5.58\% on ACC, 3.79\% on NMI, 5.79\% on ARI averagely. 
The deep learning-based methods and the GNN-based methods significantly outperform the traditional ones. 
Combined with Table~\ref{tab:dataset}, we can also observe that scCDCG achieves good results on datasets with feature dimensions of different sizes, which contain genes ranging from 2,000 to 20,670, and demonstrates higher efficiency on scRNA-seq data with high-dimension and high-sparsity.
Accordingly, in the clustering task of scRNA-seq data, scCDCG can extract intricate intercellular high-order structural information and successfully apply it to clustering methods, compared to comparative methods, outperforming the existing state-of-the-art baselines. 

\subsection{Ablation Study}
\subsubsection{Effect of each main components.} 
\begin{figure*}[t!]
\centering
\hspace{-9mm}
\subfloat[CITE\_CMBC]{
\includegraphics[width=0.24\textwidth]{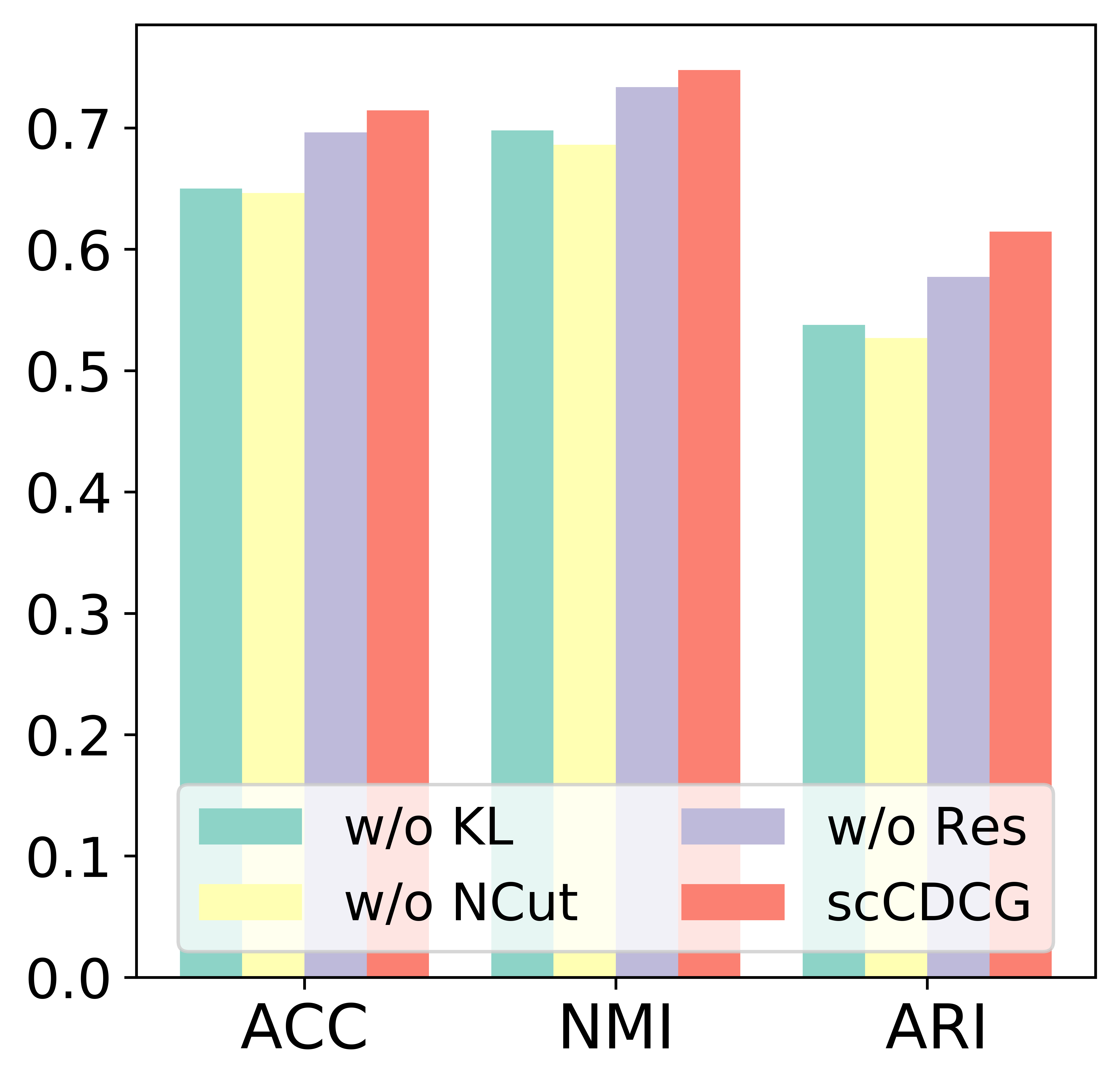}
}
\subfloat[Worm Neuron Cells]{
\includegraphics[width=0.24\textwidth]{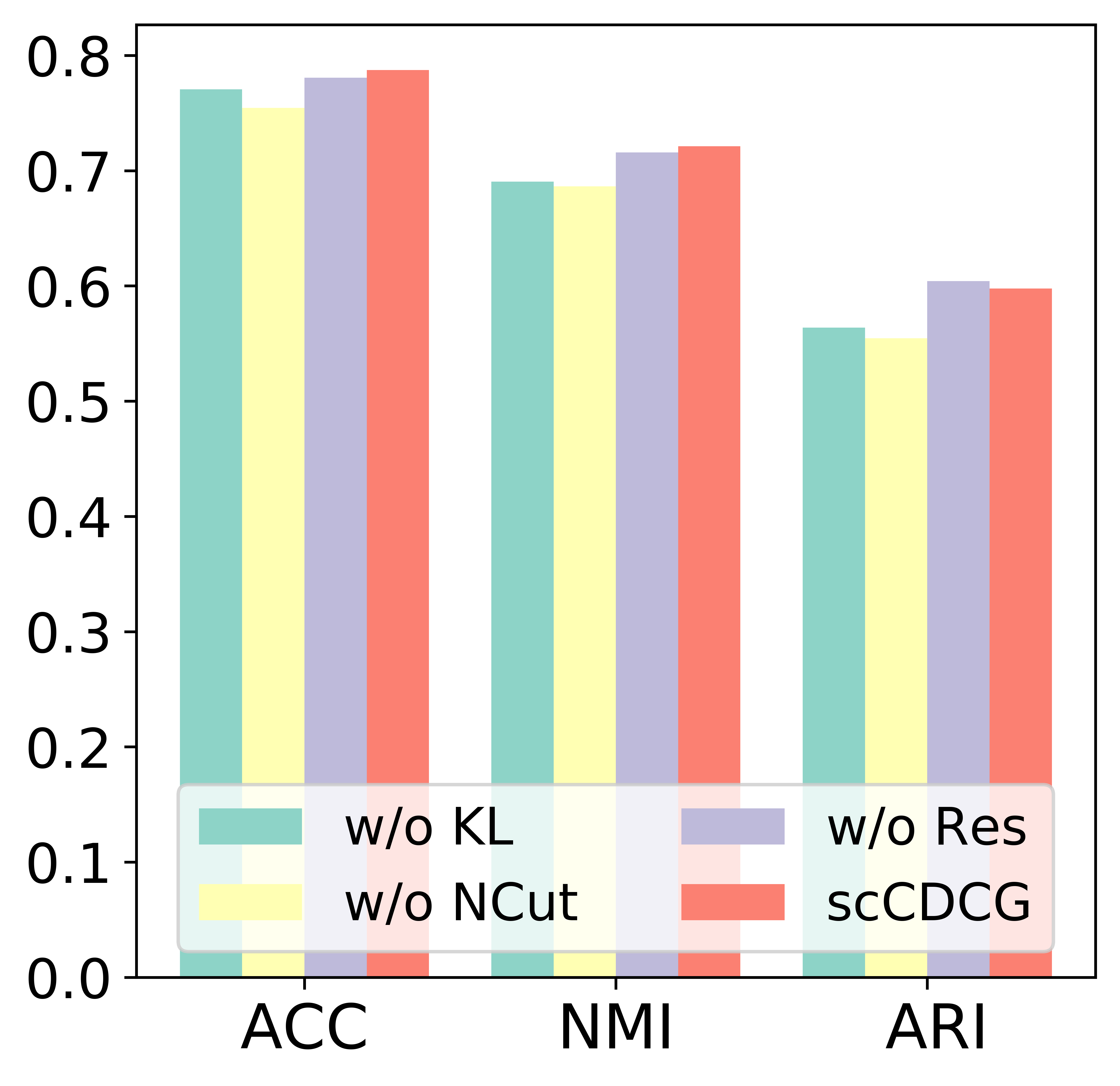}
}
\subfloat[Mouse Bladder Cells]{
\includegraphics[width=0.24\textwidth]{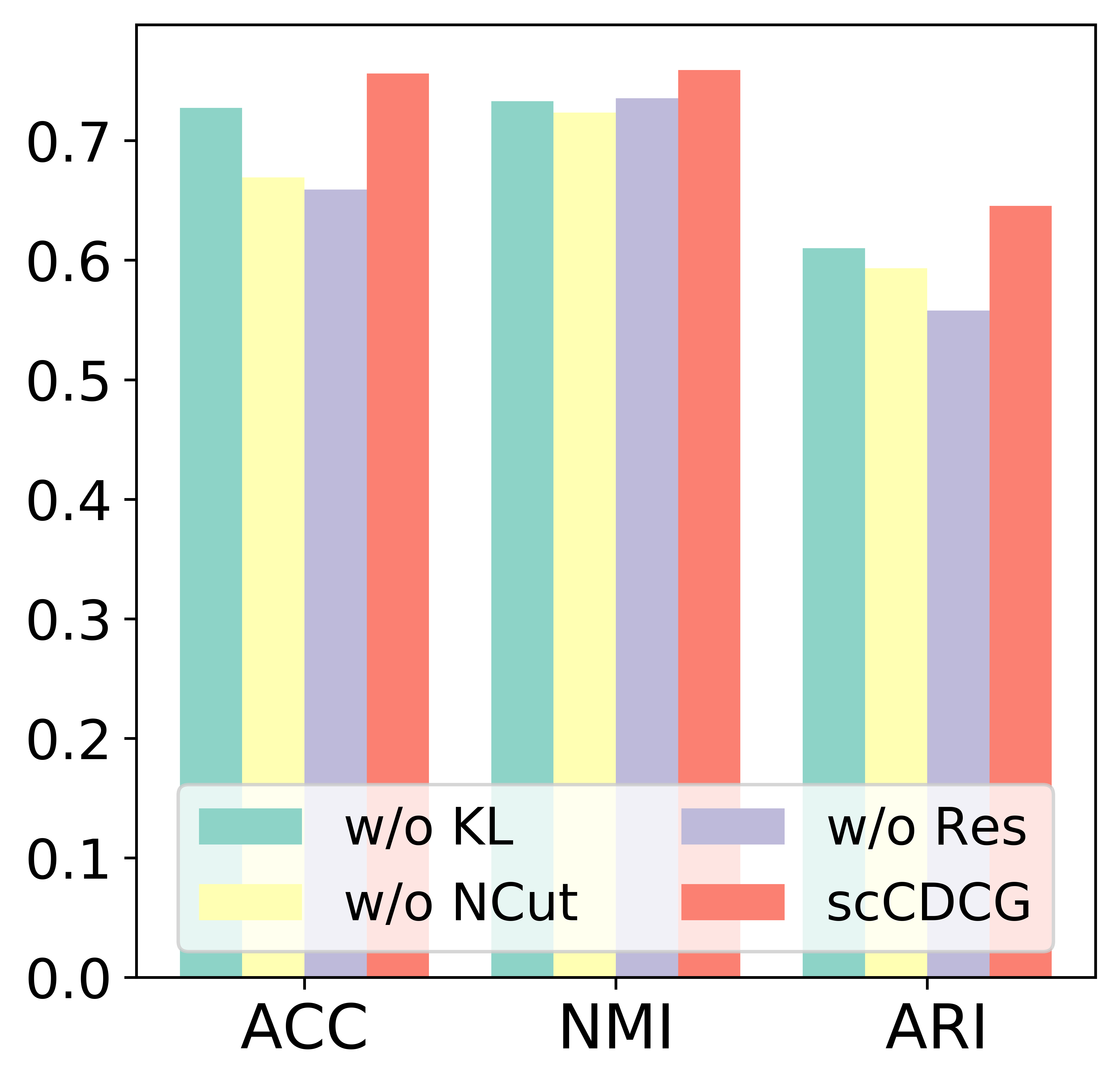}
}
\subfloat[Human Pancreas Cells 2]{
\includegraphics[width=0.24\textwidth]{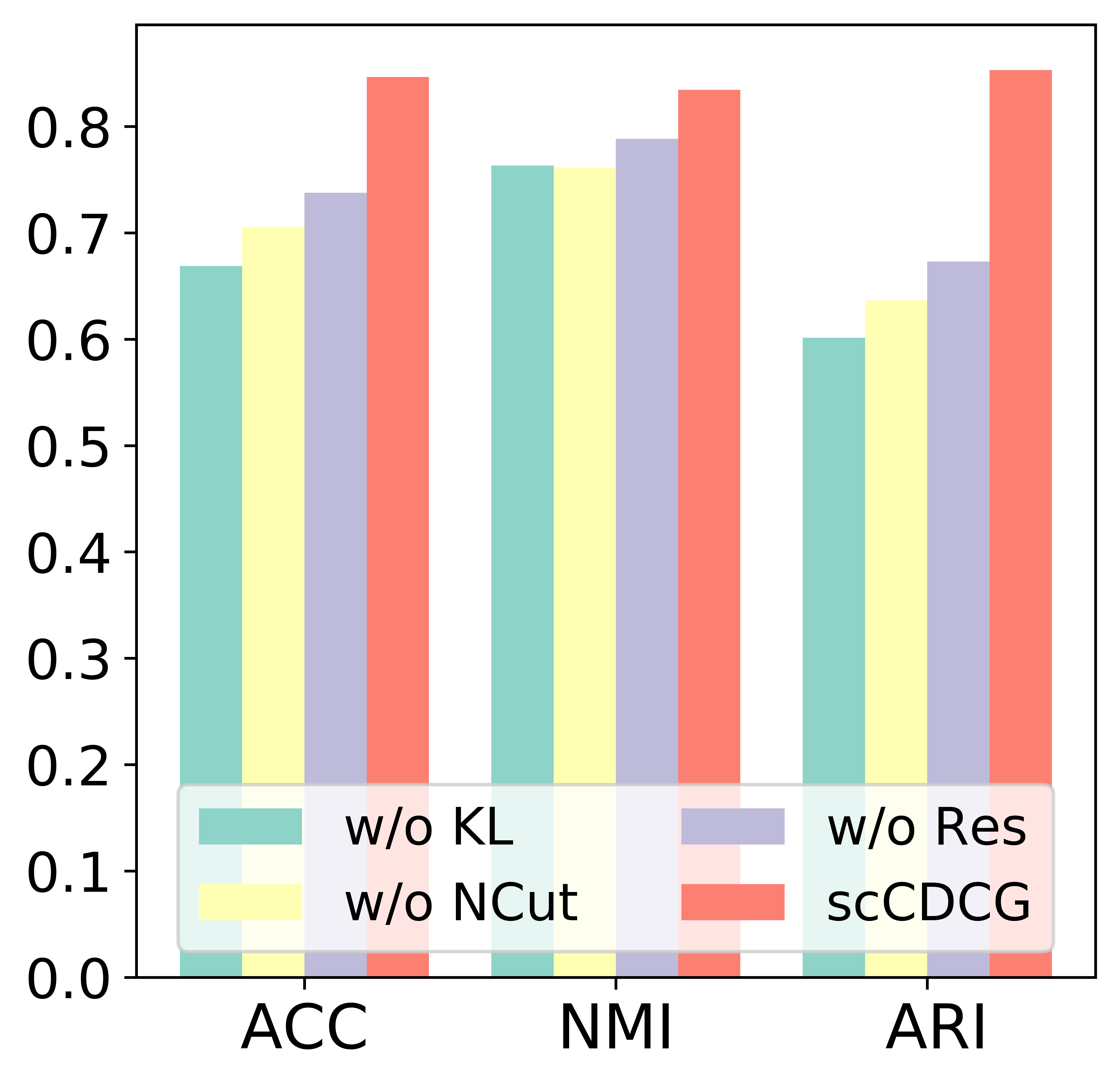}
}
\hspace{-9mm}
\caption{Ablation comparisons of all components on four datasets.}
\label{fig:ablation of all component}
\end{figure*}

We conduct ablation studies to evaluate the contributions of different components in our method. 
Here, we denote the model without NCut loss as the \emph{scCDCG-w/o NCut}. 
\emph{scCDCG-w/o KL} and \emph{scCDCG-w/o Res} denote the model without clustering loss and reconstruction loss, respectively. 
Likewise, scCDCG denotes our model that includes all components. 
Based on the results shown in Fig.~\ref{fig:ablation of all component}, we can conclude that each component of our model plays an important role and they improve the performance of scCDCG, making it more efficient.\par

\begin{figure*}[t!]
\centering
\hspace{-9mm}
\subfloat[CITE\_CMBC]{
\includegraphics[width=0.24\textwidth]{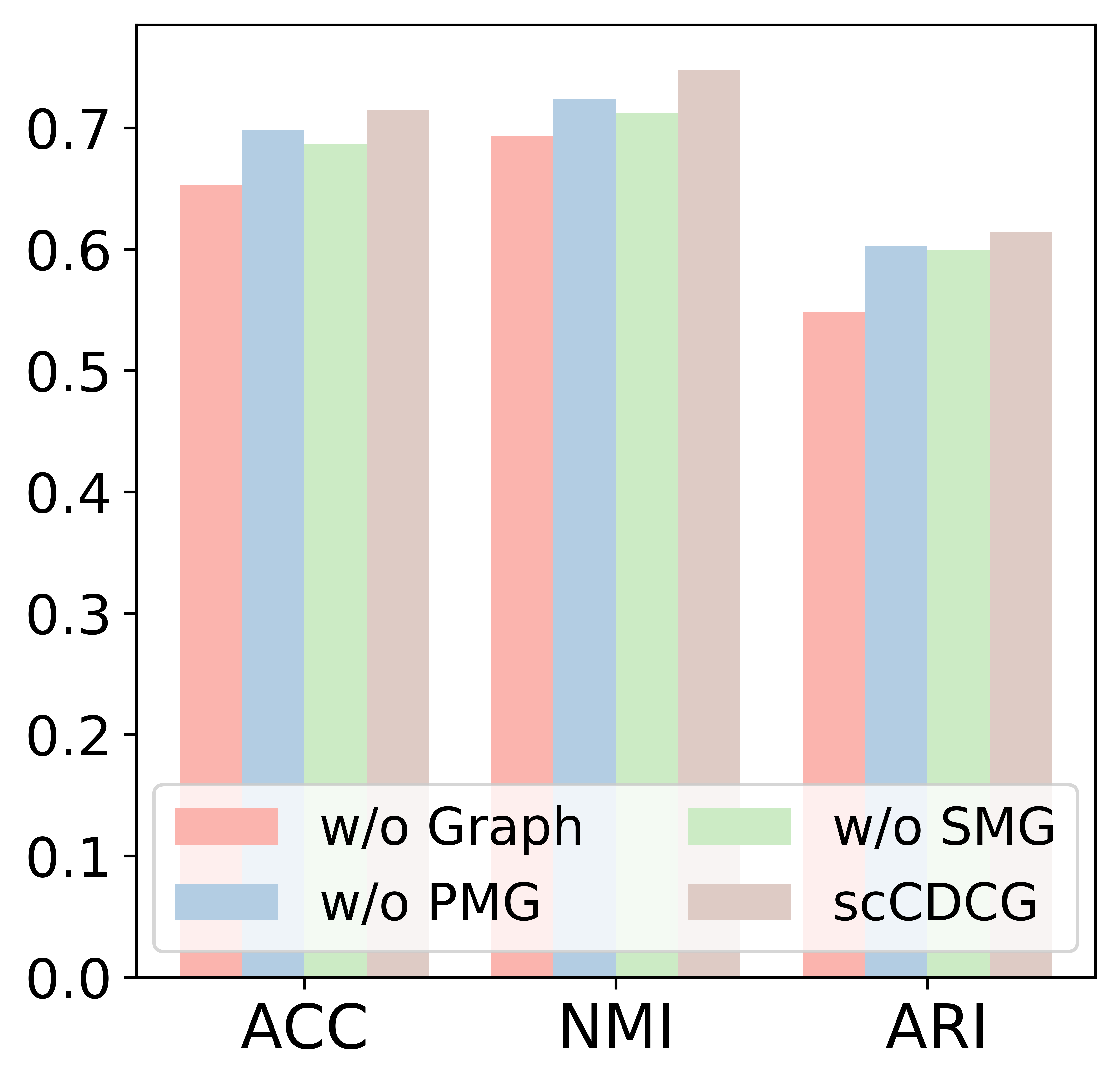}
}
\subfloat[Worm Neuron Cells]{
\includegraphics[width=0.24\textwidth]{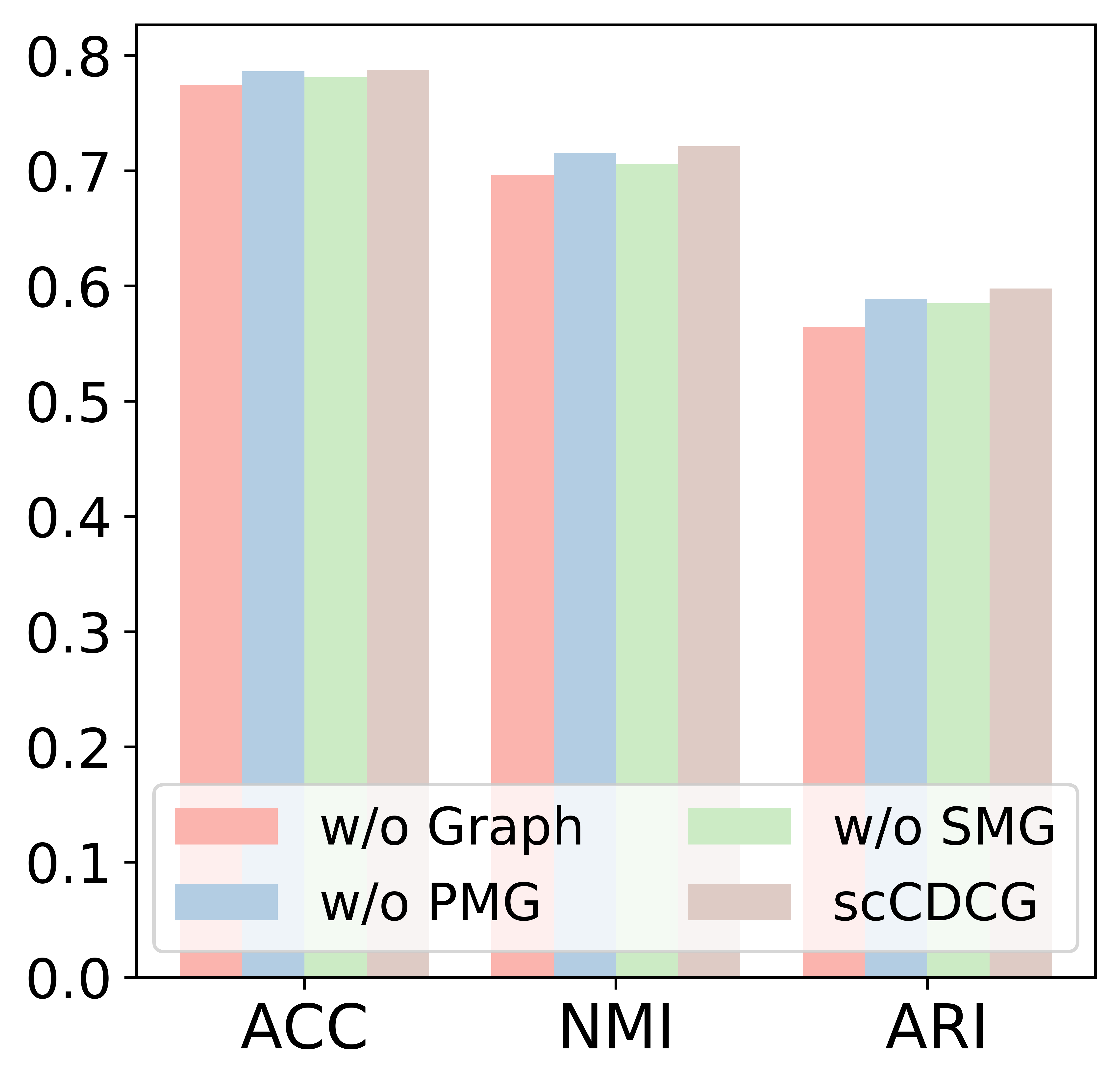}
}
\subfloat[Mouse Bladder Cells]{
\includegraphics[width=0.24\textwidth]{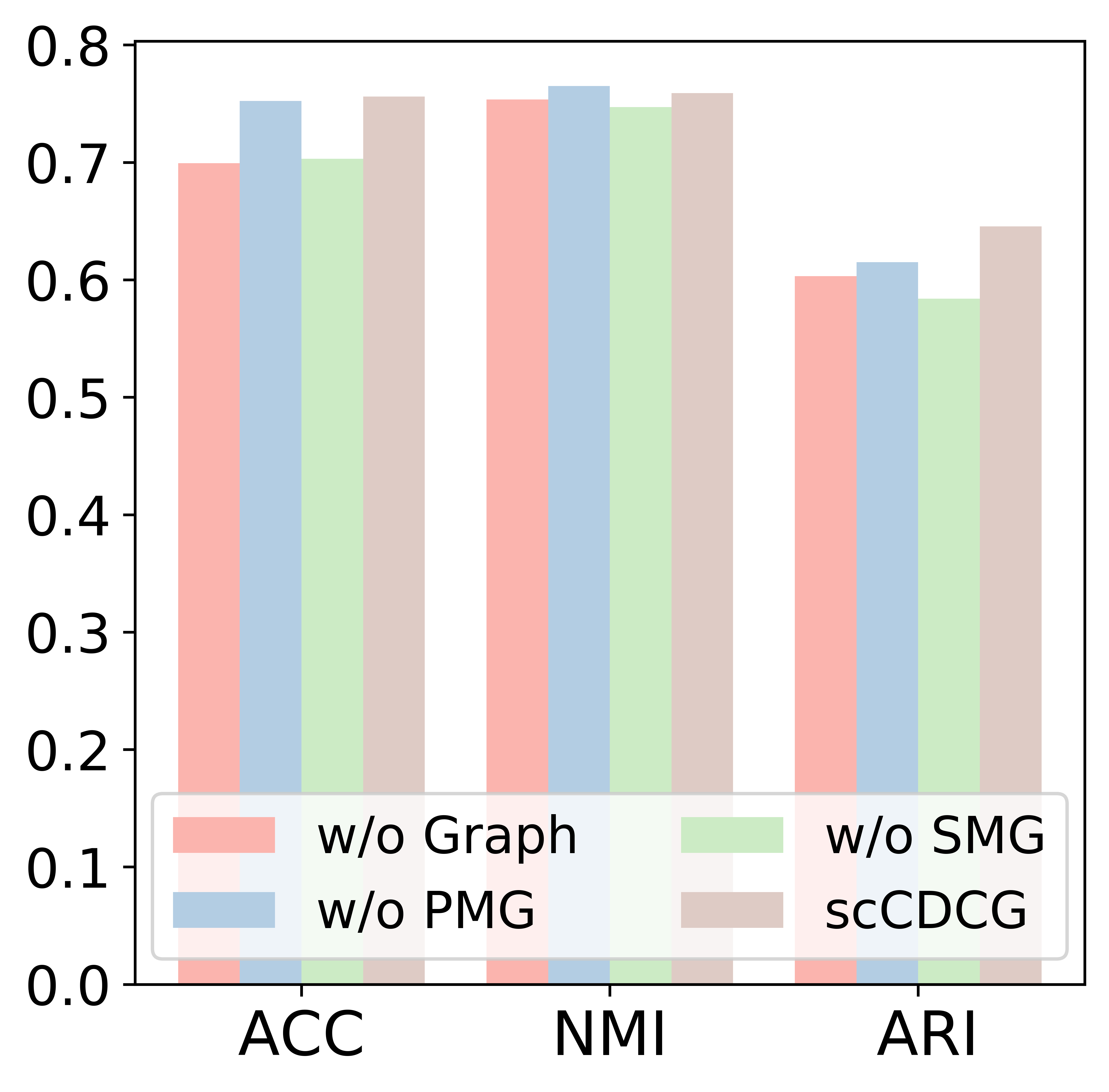}
}
\subfloat[Human Pancreas Cells 2]{
\includegraphics[width=0.24\textwidth]{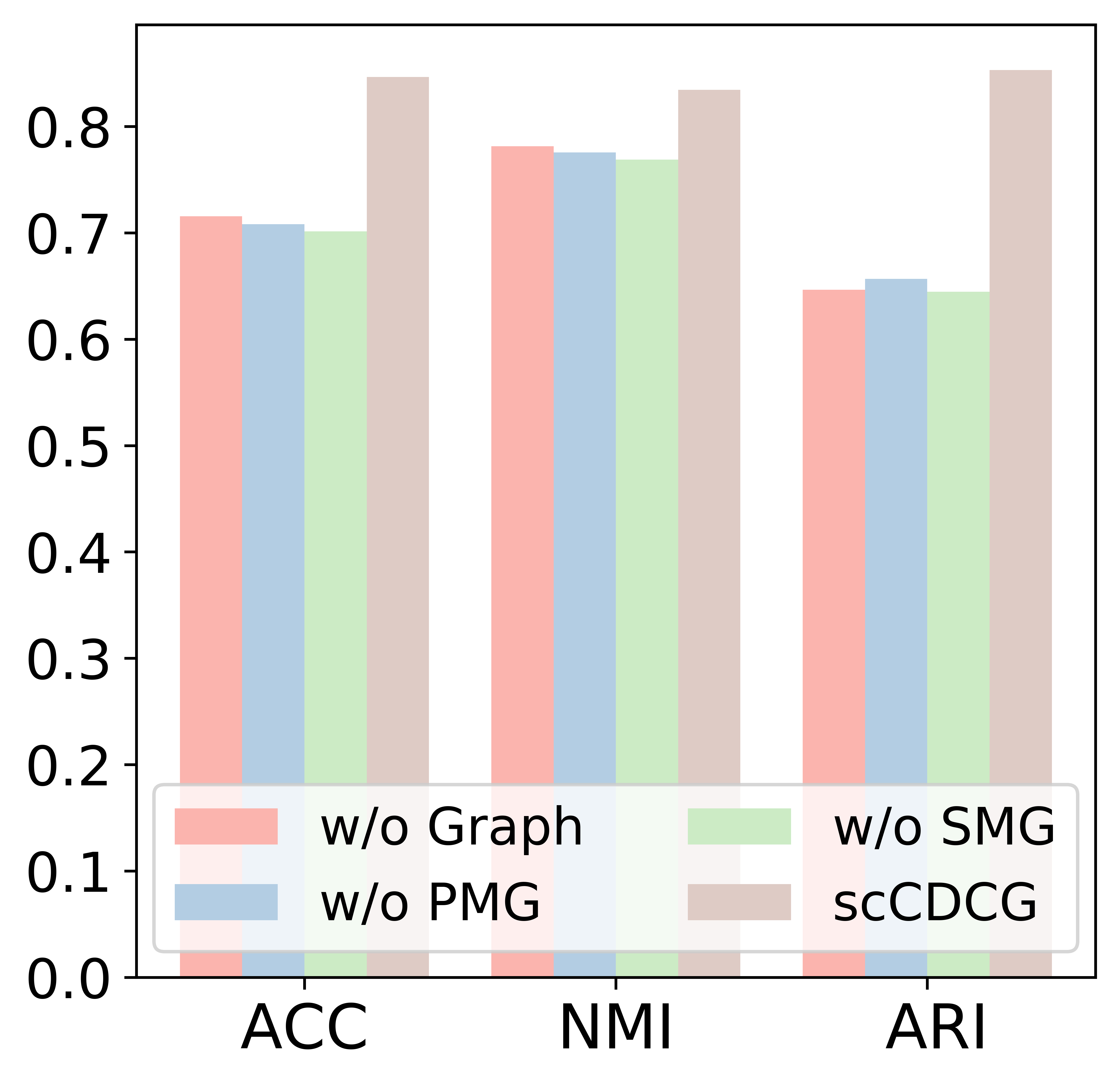}
}
\hspace{-9mm}
\caption{Effect of various graph information on model performance.}
\label{fig:ablation of graph}
\end{figure*}

\subsubsection{Effect of graph information.} To comprehensively demonstrate the significance of both the probability metrics graph (PMG) and spatial metrics graph\\(SMG), we conduct ablation studies on four distinct datasets with three metrics. 
Here, we denote the model without cut-informed graph embedding as \emph{scCDCG-w/o Graph} since it does not include probability metrics graph and spatial metrics graph.
\emph{scCDCG-w/o PMG} and \emph{scCDCG-w/o SMG} denote the model without the the probability metrics graph and spatial metrics graph, respectively. 
And scCDCG denotes our model that who contains these two graphs. 
Notably, Fig.~\ref{fig:ablation of graph} shows, each graph improves the performance compared with the baseline.
\emph{scCDCG-w/o PMG} outperforms \emph{scCDCG-w/o SMG} across all metrics, 
indicating that SMG can better capture intercellular high-order structural information between cells, with greater gains compared to SMG in the clustering task of scRNA-seq data. 
In addition, the final model scCDCG with both PMG and SMG shows superior performance compared to these baseline models, suggesting that the two graphs can interact to obtain more complex high-order intercellular structural information, crucial to account for intercellular heterogeneity in the clustering task of scRNA-seq data.\par

\subsubsection{Effect of optimal transport.} 
 We use the optimal transport theorem to guarantee the clustering process and improve the clustering assignments. To illustrate our optimal transport based self-supervised clustering module is better than the traditional approaches (e.g., DEC~\cite{xie2016unsupervised}), which adopts a clustering guided loss function to force the generated sample embeddings to have the minimum distortion against the pre-learned clustering center, we design this experiment. 
\begin{figure*}[t!]
    \begin{minipage}{0.4\linewidth}
        \centering
        \captionof{table}{Ablation study of orthogonality regularization and optimal transport.}
        \tiny 
        \renewcommand\arraystretch{1.0}
        \begin{tabular}{c|c|cc|c}
            \toprule
            \textbf{Dataset} & \textbf{Metric} & \textbf{\makecell{\emph{\shortstack{scCDCG\\{-w/o Ort}}}}} & \textbf{\makecell{\emph{\shortstack{scCDCG\\{-w/o OT}}}}} &  \textbf{\makecell{scCDCG}}\\
            \midrule
            \multirow{3}*{\textbf{\makecell{CITE-CMBC}}}
             & ACC & 56.82 & 69.45 & \textbf{71.45}\\
             & NMI & 57.99 & 71.83 & \textbf{79.34}\\
             & ARI & 45.47 & 81.26 & \textbf{81.26}\\
            \hline
            \multirow{3}*{\textbf{\makecell{Worm\\ Neuron cells}}}
             & ACC & 63.16 & 70.18 & \textbf{78.73}\\
             & NMI & 53.82 & 65.35 & \textbf{72.11}\\
             & ARI & 40.43 & 50.88 & \textbf{59.78}\\
            \hline
            \multirow{3}*{\textbf{\makecell{Mouse\\ Bladder cells}}}
             & ACC & 62.97 & 70.89 & \textbf{75.61}\\
             & NMI & 69.05 & 65.21 & \textbf{75.91}\\
             & ARI & 54.55 & 58.45 & \textbf{64.55}\\
            \hline
            \multirow{3}*{\textbf{\makecell{Human Pan\\ -creas cells 2}}}
             & ACC & 58.77 & 70.03 & \textbf{84.66}\\
             & NMI & 68.14 & 70.87 & \textbf{83.44}\\
             & ARI & 55.31 & 62.41 & \textbf{85.30}\\
             \bottomrule
        \end{tabular}
        \label{tab:expriment_ort_ot}        
    \end{minipage}%
    \hfill
    \begin{minipage}{0.47\linewidth}
        \centering
        \includegraphics[width=\linewidth, height=4cm]{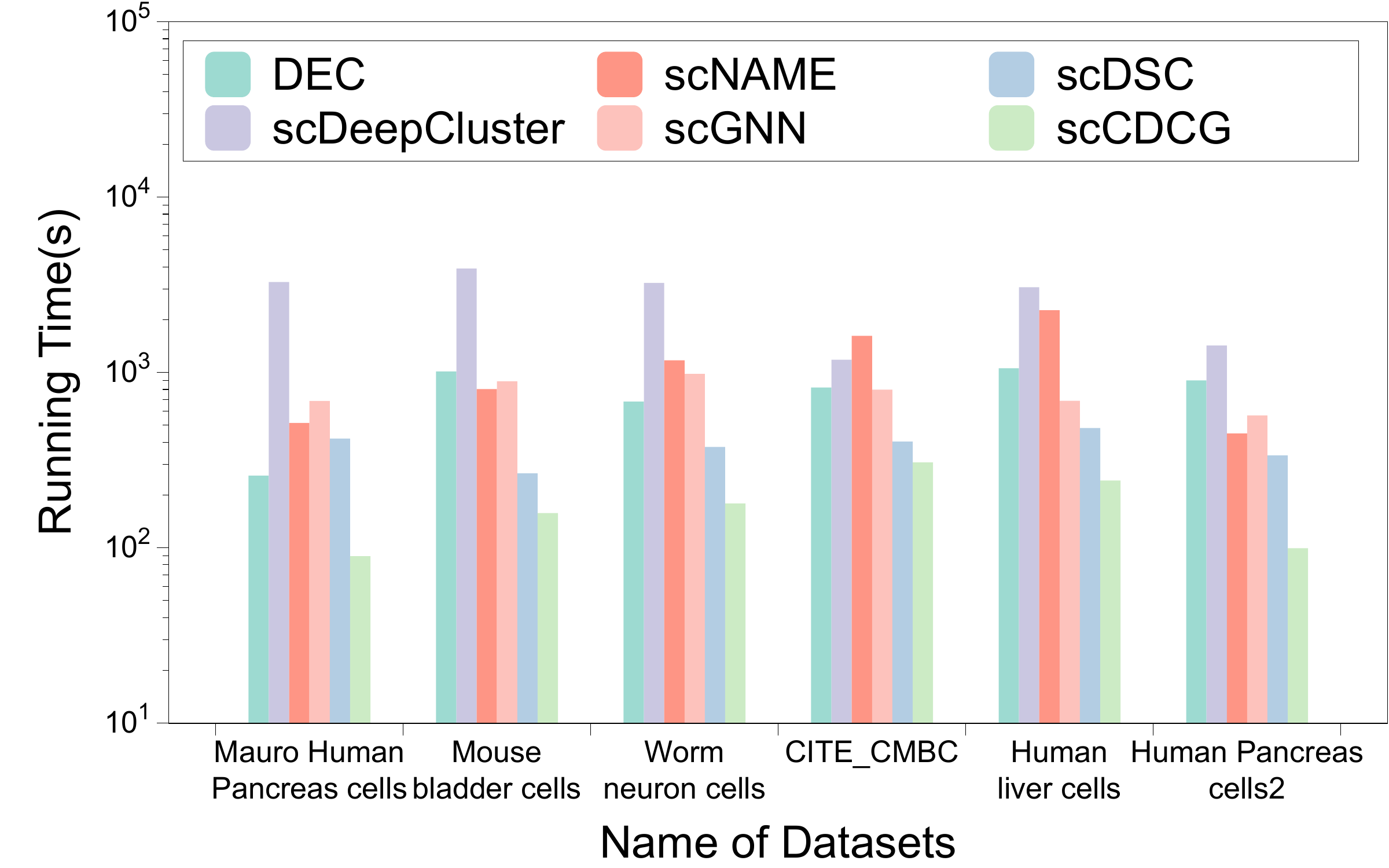}
        \caption{The running time of scCDCG and five baselines.}
        \label{fig:time_effectiveness}
    \end{minipage}
\end{figure*}
 In Table~\ref{tab:expriment_ort_ot}, \emph{scCDCG-w/o OT} denotes that the baseline does not use the proposed 
 optimal transport based self-supervised clustering module but instead uses the traditional clustering guided loss function. 
 The results show that scCDCG outperforms \emph{scCDCG-w/o OT} in terms of three metrics, indicating that introducing optimal transport can better adapt to the complex structure of scRNA-seq data, especially the two properties of high-dimension and high-sparsity.\par
\vspace{-5mm}
\subsubsection{Effect of orthogonality regularization.} 
From the perspective of minimizing normalized cut, the orthogonality of $\bm{H}$ is necessary. 
We experimentally compare our method with baselines to explore the superiority of the proposed orthogonality regularization. 
Here, \emph{scCDCG-w/o Ort} is denoted as the baseline without the proposed orthogonality regularization. 
Table~\ref{tab:expriment_ort_ot} presents the results, scCDCG consistently outperforms \emph{scCDCG-w/o Ort} across all four datasets. 
Such results indicate that the introduction of the orthogonality regularization in the graph embedding module is necessary to make the learned node embeddings sufficiently and thus avoid over-similarity and thus achieve better clustering results when processing scRNA-seq data.

\subsection{Efficiency Study}
With the increasing number of cells profiled in scRNA-seq experiments, there is an urgent need to develop analysis methods that can effectively handle large datasets. 
To evaluate the efficiency of scCDCG, we measured the runtime of five competing models (both deep learning-based and GNN-based approaches) and scCDCG on five datasets. 
As shown in Fig.~\ref{fig:time_effectiveness}, scCDCG exhibits a significant advantage in terms of time efficiency on datasets of varying sizes, while competing methods exhibit varying degrees of computational burden. This shows that scCDCG remains efficient in terms of runtime despite changes in dataset size. 
The excellent performance on large-scale datasets further validates the potential of scCDCG as a powerful tool for processing biological data of different sizes.
\begin{figure}[t!]
    \centering
    \includegraphics[width=1\linewidth]{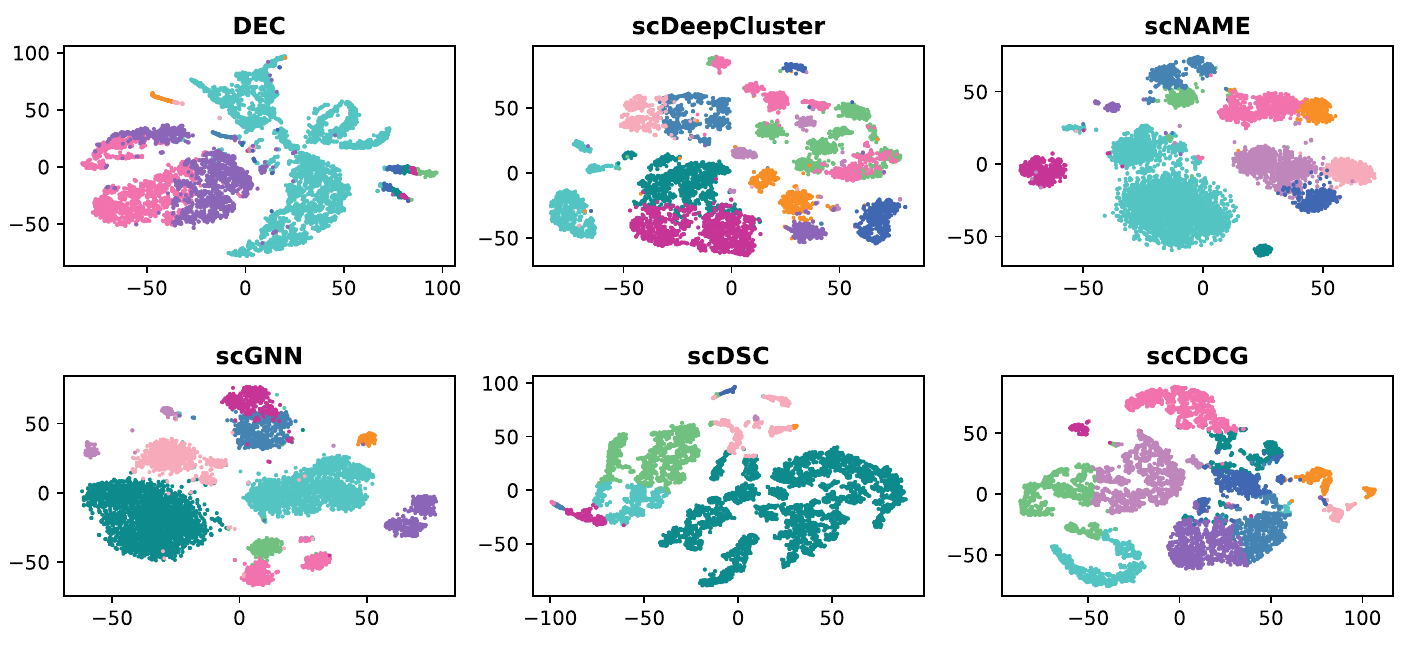}
    \caption{The visualization of scCDCG and five baselines on \emph{human liver cells.}}
    \label{fig:visualization}
\end{figure}

\subsection{Visualization}
The latent space in the proposed scCDCG model is an ideal low-dimensional embedded representation of the high-dimensional input data.
To demonstrate the effectiveness of the latent space in the scCDCG model, we employ t-SNE~\cite{hinton2008visualizing} for visualizing the final embedded points in the two-dimensional(2D) space, which are reflected in $H$. Each point represents a cell, while each color represents a predicted cell type. No method uses the true label information. 
To verify the superiority of scCDCG over competing methods, we plot the 2D space representations of all methods on \emph{human liver cells}, which contains 8,444 cells and 11 cell-types(see Fig.~\ref{fig:visualization}). scCDCG has clear boundaries between various cell subtypes, where cells of the same type are separated well, and much better than competing methods. 
We can see that the representations obtained by scCDCG are discriminative, and each cluster is compact. 
Overall, the results in Fig.~\ref{fig:visualization} show that scCDCG performs well in the discrimination of cell subtypes, which can effectively capture the intercellular high-order structural information of the scRNA-seq data.

\section{Conclusion}
In conclusion, our study introduces scCDCG, an innovative framework for the efficient and accurate clustering of single-cell RNA sequencing (scRNA-seq) data. scCDCG successfully navigates the challenges of high-dimension and high-sparsity through a synergistic combination of a graph embedding module with deep cut-informed techniques, a self-supervised learning module guided by optimal transport, and an autoencoder-based feature learning module. Our extensive evaluations on six datasets confirm scCDCG's superior performance over seven established models, marking it as a transformative tool for bioinformatics and cellular heterogeneity analysis. Looking forward, we aim to extend scCDCG's capabilities to integrate multi-omics data, enhancing its applicability in more complex biological contexts. Additionally, further exploration into the interpretability of the clustering results generated by scCDCG will be crucial for providing deeper biological insights and facilitating its adoption in clinical research settings. This future work will continue to expand the frontiers of scRNA-seq data analysis and its impact on understanding the complexities of cellular systems.




\section{Acknowledgements}
This work is partially supported by the Strategic Priority Research Program of the Chinese Academy of Sciences XDB38030300, the Informatization Plan of Chinese Academy of Sciences (CAS-WX2021SF-0101, CAS-WX2021SF-0111), the Postdoctoral Fellowship Program of CPSF (No.GZC20232736), the China Postdoctoral Science Foundation Funded Project (No.2023M743565) and the X-Compass Project.
\bibliographystyle{plain}
\bibliography{reference}
\end{document}